\documentclass[letterpaper, 10pt, conference]{ieeeconf}      

\IEEEoverridecommandlockouts                              

\usepackage{amsmath} 
\usepackage{mathtools}
\usepackage{url}
\usepackage{graphicx}
\usepackage{cite}
\usepackage{subcaption}
\usepackage{hyperref}
\usepackage{footnote}
\usepackage{booktabs}
\usepackage{adjustbox}

\title{\LARGE \bf
System Calibration of a Field Phenotyping Robot with Multiple High-Precision Profile Laser Scanners}

\author{Felix Esser$^{*}$ \quad Gereon Tombrink \quad André Cornelißen \quad Lasse Klingbeil \quad Heiner Kuhlmann
\thanks{All authors are affiliated with the Institute of Geodesy and Geoinformation at the University of Bonn, Germany. }
\thanks{$^{*}$ Corresponding author, email \href{mailto:someone@somewhere.com}{esser@igg.uni-bonn.de}.}
}

\begin{document}

\maketitle
\thispagestyle{empty}
\pagestyle{empty}

\begin{abstract}
The creation of precise and high-resolution crop point clouds in agricultural fields has become a key challenge for high-throughput phenotyping applications. This work implements a novel calibration method to calibrate the laser scanning system of an agricultural field robot consisting of two industrial-grade laser scanners used for high-precise 3D crop point cloud creation. The calibration method optimizes the transformation between the scanner origins and the robot pose by minimizing 3D point omnivariances within the point cloud. Moreover, we present a novel factor graph-based pose estimation method that fuses total station prism measurements with IMU and GNSS heading information for high-precise pose determination during calibration. The root-mean-square error of the distances to a georeferenced ground truth point cloud results in 0.8\,cm after parameter optimization. Furthermore, our results show the importance of a reference point cloud in the calibration method needed to estimate the vertical translation of the calibration. Challenges arise due to non-static parameters while the robot moves, indicated by systematic deviations to a ground truth terrestrial laser scan.
\end{abstract}
%
%
%
\section{Introduction}
%
%
Climate change and its impact on crop cultivation~\cite{asseng2015rising,asseng2019climate} require a deeper understanding of plant growth and the development of high-throughput phenotypic methods~\cite{araus2014field}. Therefore, the creation of high-resolution and accurate 3D crop models in real-field environments has become increasingly popular for deriving phenotypic traits at the scale of crop plots, individual plants, or plant organs~\cite{jay2015field, magistri2021towards,marks2022precise}. These models can be acquired using multi-view imaging or lidar (light detection and ranging) systems~\cite{weiss2011plant, esser2023quality, qiu2019field}, but usually do not provide enough precision and resolution to reconstruct single crop organs.
%
%
We therefore developed a field phenotyping robot, that is capable of generating high-resolution, georeferenced 3D point clouds in agricultural fields, using two very accurate, industrial-grade, triangulation-based laser scanners and a GNSS (Global Navigation Satellite System)/INS (Inertial Navigation System) georeferencing pose estimation unit~\cite{esserfield}. Crucial for precise reconstructions is accurately determining the transformation, composed of position and orientation, between the coordinate frames of the scanner and the pose estimation unit, which we call system calibration. 
%
%
Our sensor setup and the high accuracy requirements lead to some challenges regarding this system calibration. Existing system calibration methods are developed for large field-of-view and long-range laser scanners such as Ouster and Velodyne. Since we use high-precise triangulation profile scanners with a close measurement range of up to 1.8\,m and a small field of view of about 70\,$^\circ$ system calibration is challenging. The main contribution of this paper is a system calibration method for a scanner setup as described above. \newline
%
%
There are several approaches for calibrating laser scanner systems, depending on the actual sensor setup. A simple method is the use of external sensors like ruler measurements or taking the parameters from the system's construction plan~\cite{kaartinen2012benchmarking}. Since this method is often inaccurate, especially if the sensor origins are often not accessible, we use manual measurement to get an initial guess for the parameters only. Precise methods use overlapping scans of the environment to estimate the calibration parameters by minimizing point cloud misalignments using least-squares techniques.
Often the optimization is based on the extraction and matching of geometries such as planar surfaces, spherical targets, or cylindrical objects ~\cite{heinz2015development,chan2013multi,rieger2010boresight,yu2021automatic,ravi2018bias}.\\
The approach of \cite{heinz2015development} uses a permanently installed plane calibration field for the calibration of a full 360-degree profile laser scanner mobile mapping system. The planes are scanned multiple times from several points of view. By comparing the scans with a terrestrial laser scanner (TLS) point cloud, the calibration parameters are estimated in a least-squares adjustment by minimizing the error to the reference planes.
In \cite{rieger2010boresight} and \cite{yu2021automatic} similar methods are implemented using planes and spheres in real environments based on scans from multiple points of view.
The methods of \cite{chan2013multi} and \cite{ravi2018bias} use building facades, bridge surfaces, ground patches, light poles, and highway signs for calibration.
Other approaches use point cloud quality metrics such as eigenentropy or omnivariances of the created point clouds that are minimized to calibrate~\cite{maddern2012lost,hillemann2019automatic}.
In \cite{maddern2012lost} and \cite{elseberg2013automatic} entropy-based point cloud quality metrics are used to calibrate systems equipped with 3D and 2D laser scanners. These methods define a cost function of the eigenentropy that is minimized by optimizing the calibration parameters using least-squares techniques. The method of \cite{hillemann2019automatic} uses 3D point feature metrics such as linearity, sphericity, and omnivariance for calibration. By defining and evaluating the point feature cost function depending on the unknown parameters accurate calibration results are estimated in a least-squares adjustment.
Since all mentioned calibration techniques are developed for 2D or 3D lidar sensors characterized by large field-of-view capabilities they are unsuitable for calibration of our small-field-of-view and close-range triangulation sensors. \\
%
%
Therefore we develop a calibration method that meets these challenges. A point omnivariance is used as a point cloud quality metric in the cost function of the least-squares adjustment. Calibration parameters can be estimated for one or multiple scanners simultaneously. The vertical translation of the calibration can be estimated only using additional reference information since vertical changes do not trigger discrepancies in overlapping scans~\cite{ravi2018bias}. Therefore, we incorporate a georeferenced point cloud into the calibration method. The calibration success strongly depends on the accuracy of the pose during the calibration process and is mainly influenced by the GNSS conditions~\cite{tombrink2023trajectory}. To increase the pose accuracy and to reduce systematic errors inherent to GNSS, we mount a 360-degree prism on our robot platform, track it with a total station, and fuse the millimeter precise prism positions with the IMU data and GNSS heading angle using a factor graph-based approach. For evaluation of the calibration method, we use a mobile test calibration field and capture reference scans using a terrestrial laser scanner. \\
%
%
In summary, our contributions are:
\begin{itemize}
\item An omnivariance-based calibration method for a small-field-of-view dual laser scanning system,
\item Factor graph-based pose estimation method using IMU, 360-degree prism, and GNSS heading data,
\item Experimental validation and evaluation of our method using ground truth data
\end{itemize}
%
%
%
\section{Materials and Methods} \label{sec: materials}
\subsection{Calibration Method} \label{subsec: calibration_param_opti}
Our calibration method takes the point cloud created with the laser scan data and system poses and minimizes point omnivariance features using a least-squares adjustment to optimize the parameters. In particular, point cloud creation is performed with
\begin{equation}
    \prescript{m}{}{\mathbf{x}_i} = \prescript{m}{b}{\mathbf{t}_i}+\prescript{m}{b}{\mathbf{R}_i} \left[ \prescript{b}{s}{\mathbf{t}} + \prescript{b}{s}{\mathbf{R}}  \prescript{s}{}{\mathbf{x}_i} \right]
     \label{eq:direct georef}
\end{equation}
transforming the $i$th laser point $\prescript{s}{}{\mathbf{x}}_i = [\prescript{s}{}{x_i},0,\prescript{s}{}{z_i}]^T$ from the scanner frame $s$ into the global frame $m$. The calibration parameters are defined by the transformation between the scanner frame and pose frame, declared by the translation vector $\prescript{s}{b}{\mathbf{t}} = \left[t_x, t_y, t_z\right]$ and the rotation matrix $\prescript{s}{b}{\mathbf{R}}$ depending on the Euler angles $\alpha, \beta, \gamma$. The poses of the system are composed of $\prescript{m}{b}{\mathbf{t}_i}$ and $\prescript{m}{b}{\mathbf{R}_i}$. Here we assume the poses are precisely computed. The system calibration parameters that we aim to estimate are
\begin{align}
     \mathcal{X} = \left[t_x, t_y, t_z, \alpha, \beta, \gamma \right]
     \label{eq:param}
\end{align}
assuming that they do not change over time. When using multiple scanners the vector $\mathcal{X}$ is extended by the parameters for each scanner. Georeferencing each laser point with equation \ref{eq:direct georef} assembles the point cloud
\begin{equation}
     \mathcal{P}_s = \{\prescript{m}{~}{\mathbf{x}_1},\dotsc,\prescript{m}{~}{\mathbf{x}_q}\}
     \label{eq:point_cloud}
\end{equation}
with the number of points $q$. We use an omnivariance point feature metric to estimate $\mathcal{X}$ since it delivers both accurate and robust calibration results in contrast to other point metrics~\cite{hillemann2019automatic}. The omnivariance of a point $\sigma_i$ in $\mathcal{P}$ is computed with the eigenvalues $\lambda_i$ of its 3D tensor $N_i$~\cite{weinmann2017geometric} by
\begin{equation}
     \sigma_i({\mathbf{x}_i}, N_i) = \sqrt[3]{\lambda_{1, i} \lambda_{2, i} \lambda_{3, i}}.
     \label{eq:omnivariance_point}
\end{equation}
The 3D tensor is calculated using the neighborhood points $N_i$ around $\mathbf{x}_i$~\cite{dittrich2017analytical}. The most important aspect of the calibration method is manifested in the determination of the neighborhood points for the omnivariance $o_i({\mathbf{x}_i}, N_i)$. The point cloud $\mathbf{P}$ is extended by an accurate reference point cloud of the environment $\mathcal{P}_g$. Afterward, the neighborhood points based on the point cloud $\mathcal{P} = \{ \mathcal{P}_s, \mathcal{P}_g \}$ are determined.
When calibrating $n$ scanners at the same time, the point cloud for omnivariance calculation results in
\begin{equation}
     \mathcal{P}' = \{ \mathcal{P}_{s,1}, \dotsc, \mathcal{P}_{s,n}, \mathcal{P}_g \}.
     \label{eq:point_cloud_multiple}
\end{equation}
The cost function sums up all squared point omnivariances of the $n$ scanners to be calibrated and is summarized by
\begin{align}
    \mathcal{C} = \sum_{i=1}^{q_1} \sigma_i\left( \mathbf{x}_i, N_i \right)^2 + \dotsc + \sum_{i=1}^{q_n} \sigma_i\left( \mathbf{x}_i, N_i \right)^2, \mathbf{x}_i \in \mathcal{P}'
    \label{eq: cost function}
\end{align}
Since $\mathcal{C}$ describes a nonlinear function of the calibration parameters, the robot poses, and laser profiles, a linearization is required. This is implemented by estimating the Jacobians of \ref{eq: cost function} with respect to the parameters $\mathcal{X}$. Gradient-based methods are suitable for this linearization, for example. Parameter updates are estimated in an iterated least-squares adjustment until the updates are significantly small. 
For optimization, we follow the multi-scale approach of~\cite{hillemann2019automatic} since it yields promising results in terms of robustness and accuracy. The method down-samples the point cloud $\mathcal{P}'$ using a voxel grid filter. After parameter convergence on this first, rough voxel grid, the grid is refined and the optimization is repeated. The calculated parameters on each scale are used as an initial guess for the next scale. This process is iterated until the last scale is reached, giving the optimized parameters.
\subsection{Agricultural Field Robot} \label{subsec: robot_platform}
\begin{figure}[ht]
      \centering
      \includegraphics[width=1.0\linewidth]{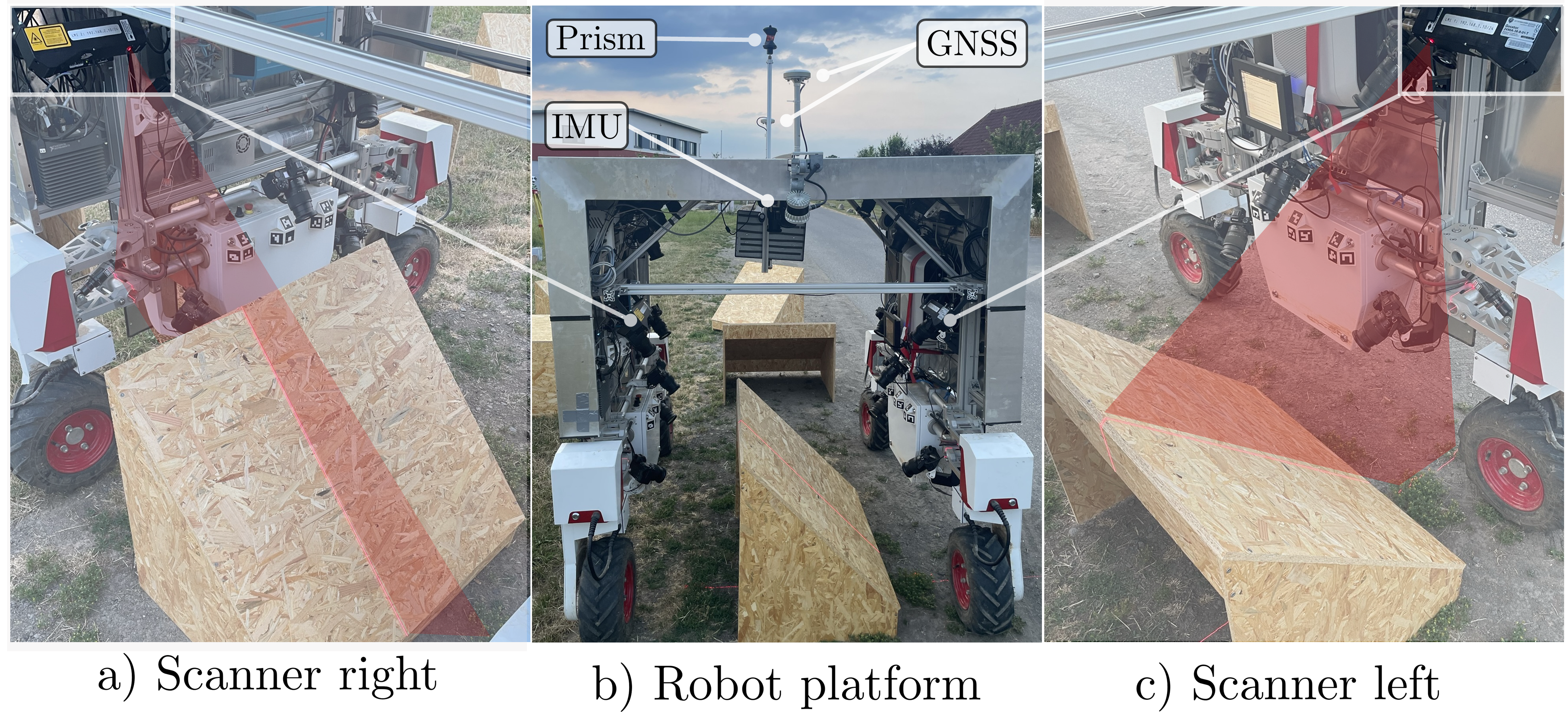}
      \caption{a) Right laser triangulation scanner. b) For pose estimation: front and back GNSS antenna, IMU, and 360-degree prism. c) Left laser triangulation scanner.}
      \label{fig:calibration field}
\end{figure}

%
%
The agricultural field robot platform designed for crop phenotyping is shown in figure \ref{fig:calibration field}a and has the dimensions of about 1.8\,m$\times$2\,m$\times$2\,m in length, width, and height. The platform is equipped with four electric motors powered by a 48\,V lithium battery and is controlled via a remote control to drive in fields. The lightweight U-shaped aluminum enclosure allows a safe mounting of the sensors and computers shielded from external environmental influences like wind, sunlight, or rain. To reduce the effects of vibrations on the robot and its sensors while driving it is extended with suspensions on the wheels.
%
%
\subsubsection{Georeferencing Sensors} \label{subsec: georef}
The georeferencing system includes an INS (Inertial Navigation System) SBG ellipse D with an internal Multi-GNSS (Global Navigation Satellite System) receiver and two antennas for position and heading determination attached to the roof in the back and front. The GNSS receiver estimates the position and heading of the robot using RTK (Real-Time Kinematic) using base station data at a measurement rate of 5\,Hz. The INS also contains an inertial measurement unit (IMU) measuring 3D accelerations and angular velocities at a rate of 100\,Hz.
For precise pose estimation during calibration drives we mount a 360-degree prism (Leica GRZ4) on the robot roof at an aluminum profile in a height of about 0.5\,m to the robot roof, see figure \ref{fig:calibration field}b. The prism of the robot is tracked using an automated and motorized Leica TS60 Total Station (TS) giving prism positions with an accuracy of about 5\,mm at a rate of about 7\,Hz. For the pose estimation approach in section \ref{subsec: pose_estimation}, the TS measurements must be time-stamped with the IMU and GNSS data. The prism positions are synchronized using a Raspberry Pi 3 and a u-blox EVU-6T GNSS receiver with an external antenna. The system time of the Pi is synchronized using the PPS (Pulse Per Second) pulse coming from the Ublox receiver. The automated prism tracking using the motorized total station is realized with a script implementation that uses the Leica Geocom interface. The script runs on the Pi, reads the prism measurement from the interface, and stores it with the PPS synchronized system time of the Pi.
%
%
\subsubsection{Laser Triangulation Scanner} \label{subsec: laserscanning_system}
The laser scanning system of the field robot consists of two laser profile triangulation sensors (LMI Gocator 2490) attached to the inner enclosure of the robot, see figure \ref{fig:calibration field}. The scanners measure up to 1920 points per profile depending on the reflectivity of the scanned surfaces. The scanner measures 200 laser profiles per second resulting in an along-track point density of 0.5\,mm at a driving speed of 0.1\,m/s. According to the manufacturing sensor manual, the point repeatability is valued at 12\,$\mu$ under lab conditions. The laser fan field-of-view is about 70$^\circ$ at a maximum measurement range of about 1.8\,m. For precise time synchronization the laser lines of the LMI scanners are timestamped using the PPS signal coming from the SBGs GNSS receiver on an industrial-grade computer.
\subsection{Pose Estimation} \label{subsec: pose_estimation}
As it can be seen in equation~\ref{eq:direct georef}, the accuracy of the pose significantly influences the accuracy of each transformed point. Therefore, it also strongly influences the accuracy of the calibration parameters during the calibration process. For accurate pose estimation of the robot platform during calibration, we use a factor graph-based approach fusing the sensor data of the IMU, the GNSS heading, and TS prism measurements.
\subsubsection{Factor Graph Optimization} \label{subsec: factor_graph}
\begin{figure}[ht]
      \centering
      \includegraphics[width=1.0\linewidth]{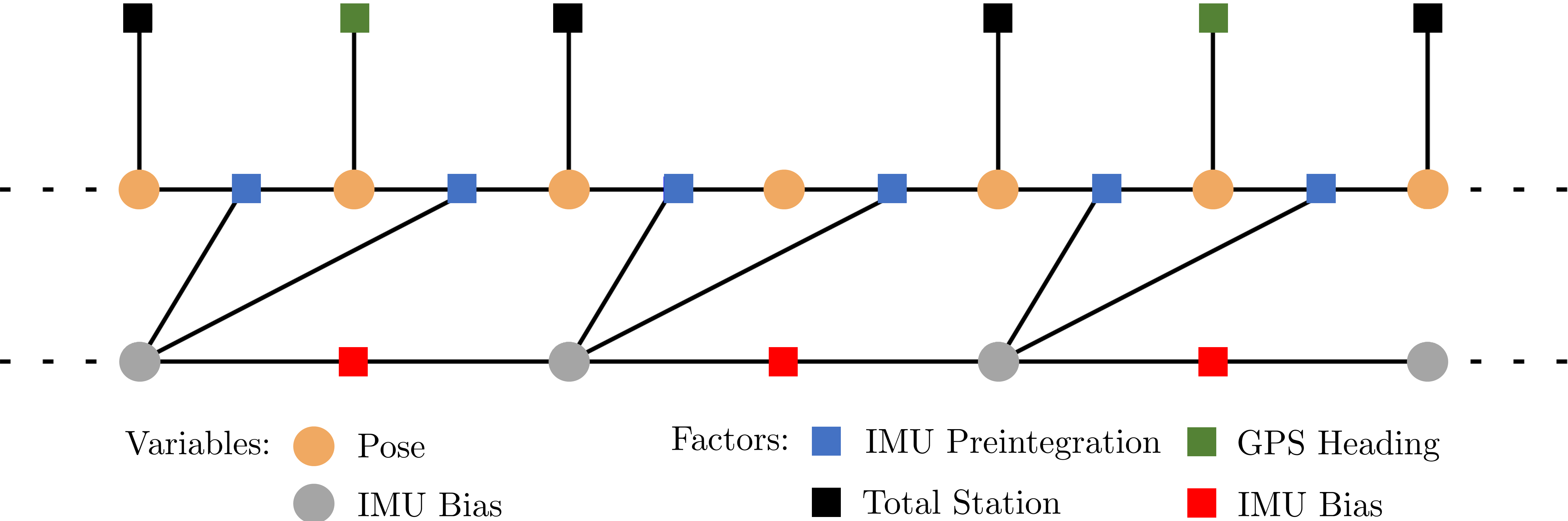}
      \caption{Factor graph of the pose estimation.}
      \label{fig:factor_graph}
   \end{figure}

The pose estimation framework uses a factor graph-based approach~\cite{dellaert2012factor} implemented in the gtsam library\footnote{\url{https://gtsam.org/}}. The graph representation of the pose estimation framework is shown in figure \ref{fig:factor_graph}. Circles mark nodes representing unknown variables for the robot poses and the IMU bias. Squares inside the graph are factors, defining non-linear error functions depending on the pose and IMU bias variables and the sensor measurements coming from the total station, GNSS, and IMU.
%
%
The pose variable consists of the translation, velocity, and attitude of the robot summarized by the set $\mathbf{P} = \{\mathbf{t}, \mathbf{v}, \mathbf{q}\}$. The IMU bias variables are composed of the acceleration and angular velocity bias of all three measurement axes.
%
%
The factor graph consists of four factors that are visualized in figure \ref{fig:factor_graph}. The IMU preintegration factor inputs the acceleration and angular velocities measured by the IMU and involves two pose variable nodes. The IMU bias factor models a random walk process, connects two IMU bias variable nodes, and describes the temporal development of the biases. More information on the IMU preintegration and with bias estimation gives \cite{forster2016manifold,carlone2014eliminating,lupton2011visual}. The total station and the GNSS heading factor are so-called unary factors since they depend on a single pose variable only. The input measurements are the measured prism position and GNSS heading respectively. \\
%
%
The factor graph implementation introduces a pose variable with every IMU measurement resulting in a rate of 100 poses per second. The TS prim positions are included with a standard deviation of 5\,mm at a rate of about 7\,Hz. The GNSS heading angle is included with a standard deviation of about 0.7\,$^\circ$, coming from the receiver at a measurement rate of 5\,Hz. The IMU bias variables are estimated at the rate of the TS prism measurements. For optimization of the variables, we use the iSAM2 algorithm~\cite{kaess2012isam2, kaess2011isam2} in post-processing to optimize the factor graph delivering both precise and smooth poses of the robot platform. These poses are used in equation \ref{eq:direct georef} of the calibration method. \\
Please note that this technique could also be used for pose estimation during normal measurements in agricultural fields, but it is an offline algorithm, that requires additional total station measurements that are elaborate in setup. Therefore it is only applied for calibration datasets.
\subsection{Calibration Setup and Experiments}
%
%
\subsubsection{Calibration Field} \label{subsec:calib_field}
A system calibration requires observations of certain structures in the environment, which allow an estimation of the calibration parameters~\cite{ravi2018bias}. For the robot, we design a calibration field consisting of five planes of different orientations as shown in figure~\ref{fig:calib_field}. It has a size of about 7.5\,m times 6\,m. The attitude of the planes varies between 0 and 55\,$^\circ$. To fit under the robot, the spatial dimensions of the planes are about 1\,m$\times$1\,m$\times$0.5\,m. We choose plane objects since shifts and tilts coming from an inaccurate calibration can easily be discovered in the later evaluation of our method.
%
%
\subsubsection{TLS Point Cloud} \label{subsec:georef point cloud}
\begin{figure}[ht]
      \centering
      \includegraphics[width=1.0\linewidth]{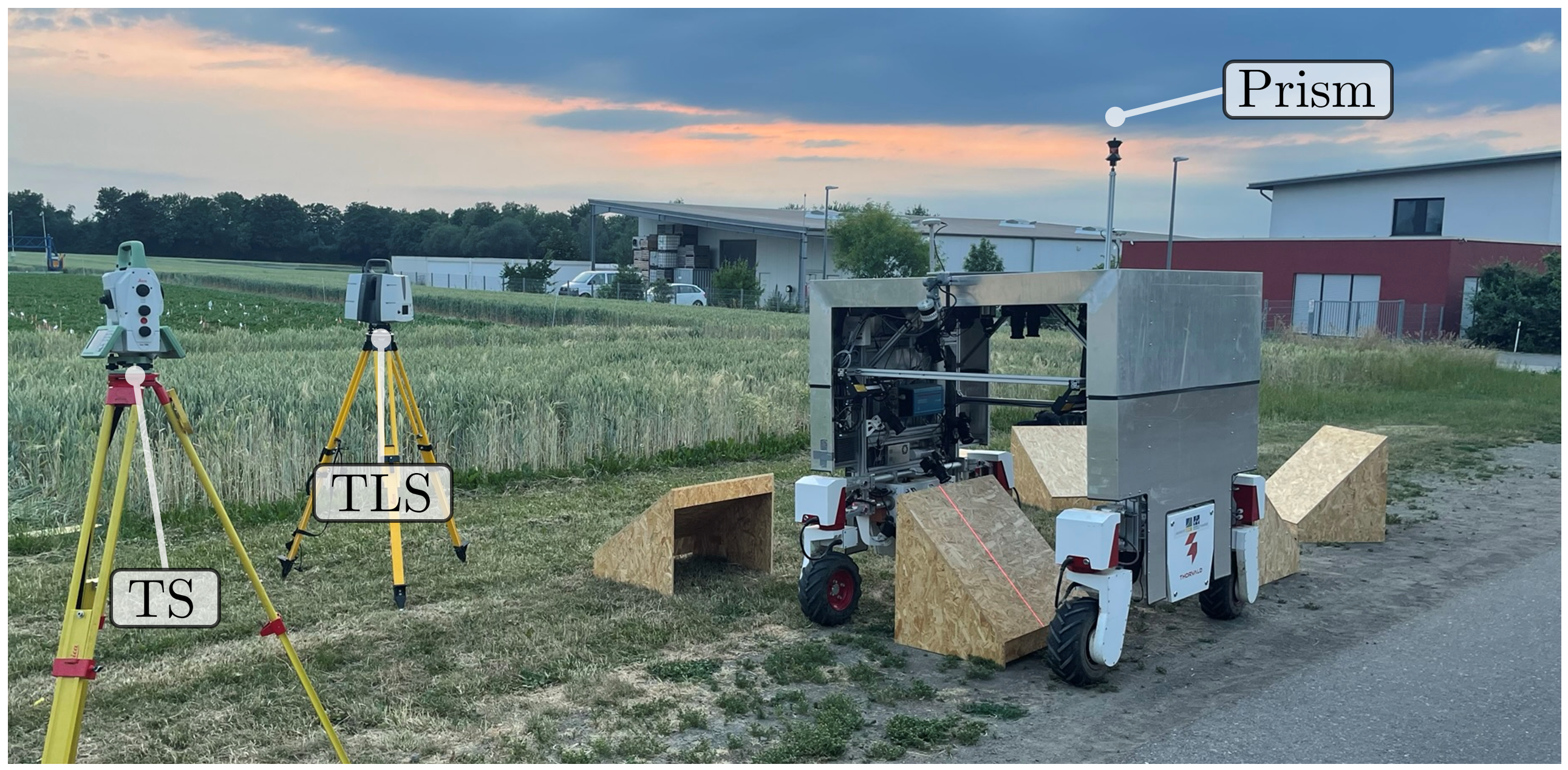}   
      \caption{Calibration test field used to evaluate our calibration approach. The prism is tracked with the Total Station (TS) for precise pose estimation. The Terrestrial Laser Scanner (TLS) is used to create a georeferenced point cloud for calibration and evaluation.}
      \label{fig:calib_field}
\end{figure}

To create a georeferenced point cloud for the calibration method a Leica P50 TLS is used to scan the plane setup of figure \ref{fig:calib_field} from four viewpoints. The single scans are registered using seven TLS targets around the calibration field. Finally, the point cloud is georeferenced with four targets whose coordinates are known with millimeter accuracy based on a network adjustment using Leica TS60 total station measurements. The resulting point cloud is sub-sampled to spatial point-to-point distances of 1\,mm to accelerate the omnivariance estimation.
\subsubsection{Datasets \& Calibration}
\begin{figure*}[ht]
      \centering
      \includegraphics[width=1.0\linewidth]{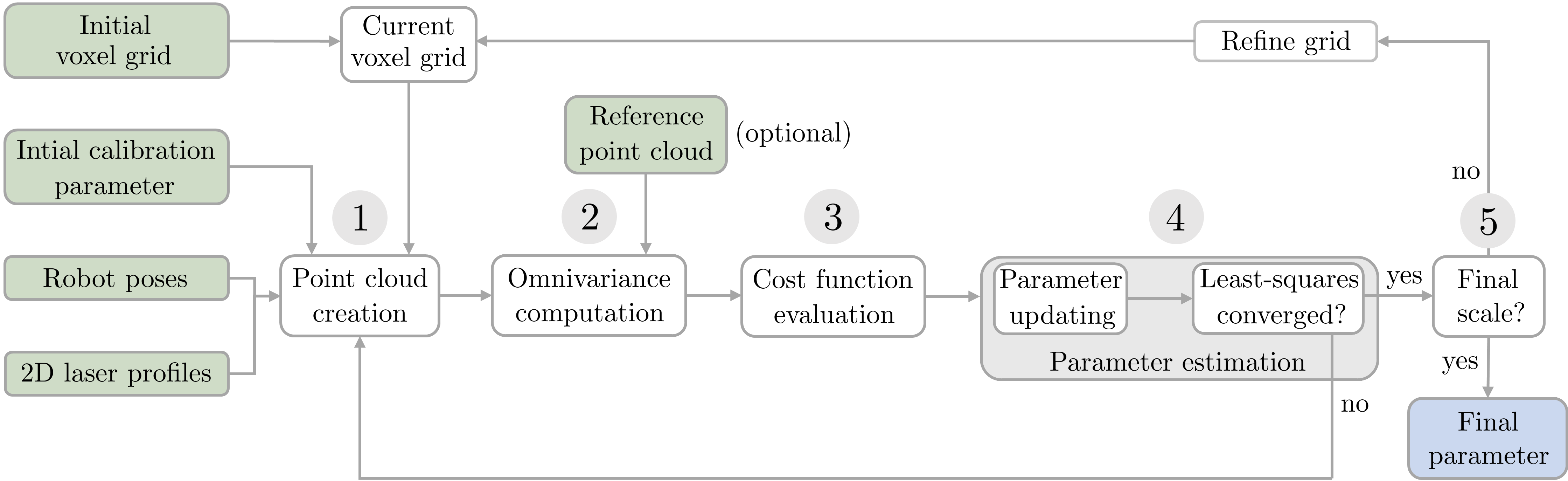}    
      \caption{Pipeline of our calibration method. 1) Point cloud creation with 2D laser profiles, initial calibration parameters, and robot poses. 2) Omnivariance computation with point cloud from 1 and reference point cloud (optional). 3) Evaluation of the cost function. 4) Iterative least-squares optimization. 5) Estimation on multiple scales.}
      \label{fig:system_calibration_pipeline}
\end{figure*}

%
%
For data capturing the robot is remotely steered through the calibration field. At the beginning of the drive, the TS measurement script on the Raspberry Pi is started, storing the synchronized prism positions. Simultaneously, the GNSS heading and IMU data are recorded on the robot computer. In post-processing, all captured data are fused using the factor graph-based approach described in section \ref{subsec: pose_estimation}. When scanning the planes in figure \ref{fig:calib_field} the driving speed of the robot is reduced to 10\,cm/s to increase the spatial resolution along the driving direction. Reasonable parameter estimates need objects to be scanned from multiple points of view. For that reason, the robot trajectories are planned in a way to scan each plane twice from different viewpoints. \\
%
%
To analyze the consistency of the calibration estimates in later evaluation, three datasets are recorded, including different trajectories in the calibration field. For each dataset, we capture the accelerations and angular velocity of the IMU, the GNSS heading angle, the TS prism positions, and the laser profiles of the triangulation scanners. Figure~\ref{fig:system_calibration_pipeline} shows the calibration pipeline. It inputs the estimated robot poses, coming from the factor graph optimization, the laser profiles of both scanners, the initial calibration parameters, the reference point cloud, and the initial voxel grid size used on the first scale. The inputs are highlighted in green. Including the reference point cloud is optional since we evaluate its importance in the later evaluation. Since no online calibration is needed in our applications, the pipeline calibration runs offline after taking the measurements. \\
%
%
The optimization is performed on six scales with the voxel grid sizes of 5, 4, 3, 2, 1, and 0.5\,cm. The finest scale is 0.5\,cm since no significant parameter changes were detected on finer scales. The initial translation of the calibration is determined with ruler measurements. Approximations for the rotation to the robot pose frame are taken from a construction plan. On each scale, the parameters are optimized until the updates on the parameters are significantly small. The number of iterations needed until convergence on each scale varies between 50 and 450 approximately.
%
%
\section{Results and Discussion} \label{sec: experiments_results}
\begin{figure}[ht]
      \centering
      \includegraphics[width=1.0\linewidth]{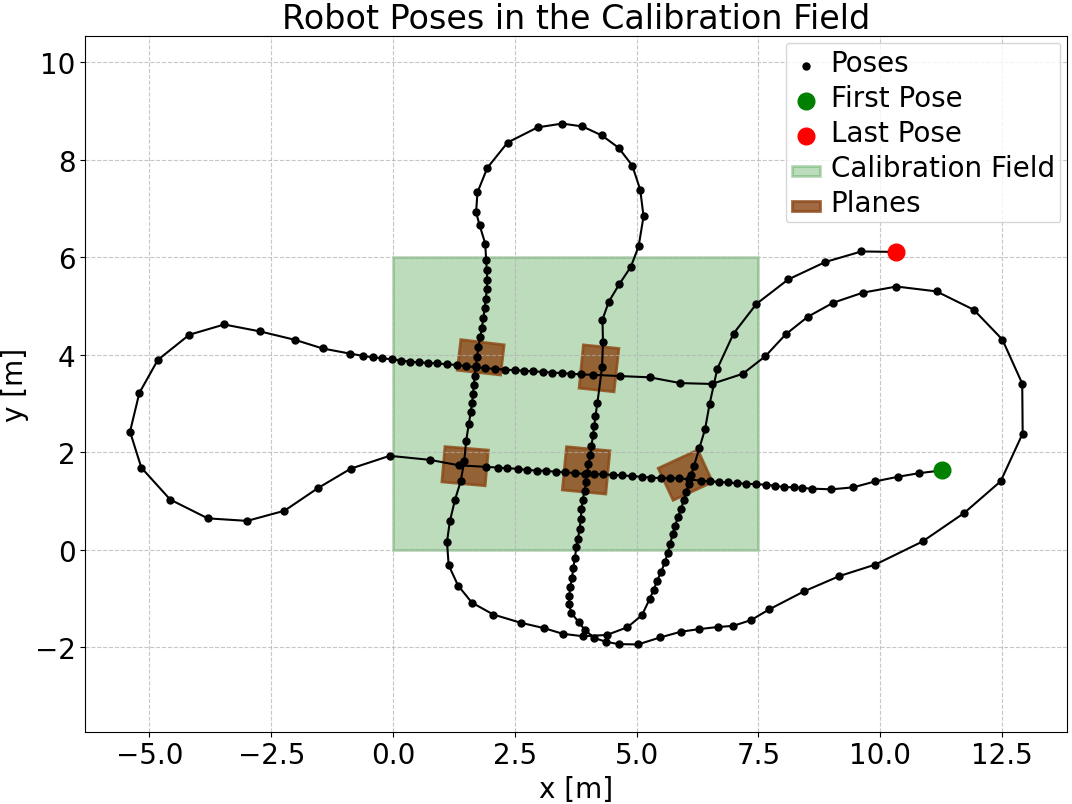}   
      \caption{Poses of the robot estimated by our graph-based pose optimization approach using TS, IMU, and GNSS heading data as introduced in \ref{subsec: pose_estimation}. The brown rectangles represent the planes as shown in figure \ref{fig:calib_field}. }
      \label{fig:traj_robot_calib_field}
\end{figure}

%
%
Figure~\ref{fig:traj_robot_calib_field} shows the estimated robot poses for the first dataset. The calibration field is highlighted in green, the planes are shown in brown. The first and last robot poses are marked in green and red, and the remaining poses are black circles connected by lines. The number of poses is reduced by a factor of 50 to improve the visualization.\\ We do not have the chance to evaluate the pose estimation separately as there is no independent pose information available that has a higher accuracy. Therefore, only the quality of the result of the calibration procedure is evaluated, containing also the pose accuracy.

We evaluate the calibration result by comparing the resulting point cloud after parameter optimization with the reference point cloud coming from the georeferenced terrestrial laser scan. This is done by using the multi-scale model-to-model cloud comparison (M3C2)~\cite{lague2013accurate}. It compares two point clouds by first determining point correspondences using surface normals and computes signed point distances afterward. Although it might be questionable, if a single number can describe the difference between two clouds in an appropriate way, we calculate the root mean square error of all M3C2 distances between the two clouds. We consider this value as the metric for the quality of the reconstructed point cloud and therefore of the calibration procedure as a whole. Below we analyze the influence of integrating the reference scan into the calibration, the consistency between the two scanners, and the transferability between the datasets captured.
%
%
\subsection{Influence of the reference scan in the calibration} \label{subsec: infleuncereference}
\begin{figure*}[!htb]
    \centering
    \label{fig:m3c2}
    \begin{subfigure}{0.32\linewidth}
        \centering
        \includegraphics[width=\linewidth]{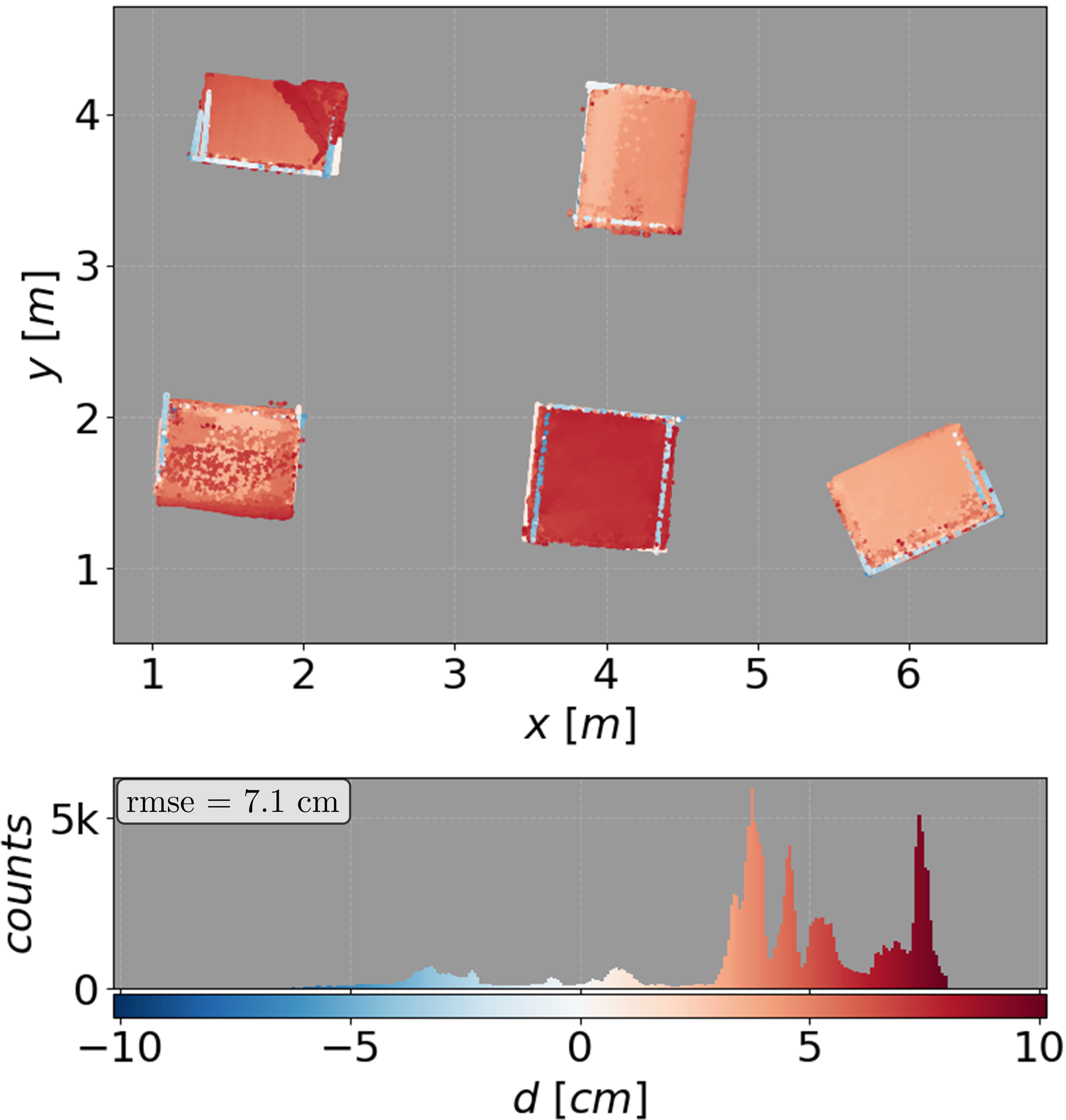}
        \caption{Initial calibration}
        \label{subfig:bild1}
    \end{subfigure}
    \begin{subfigure}{0.32\linewidth}
        \centering
        \includegraphics[width=\linewidth]{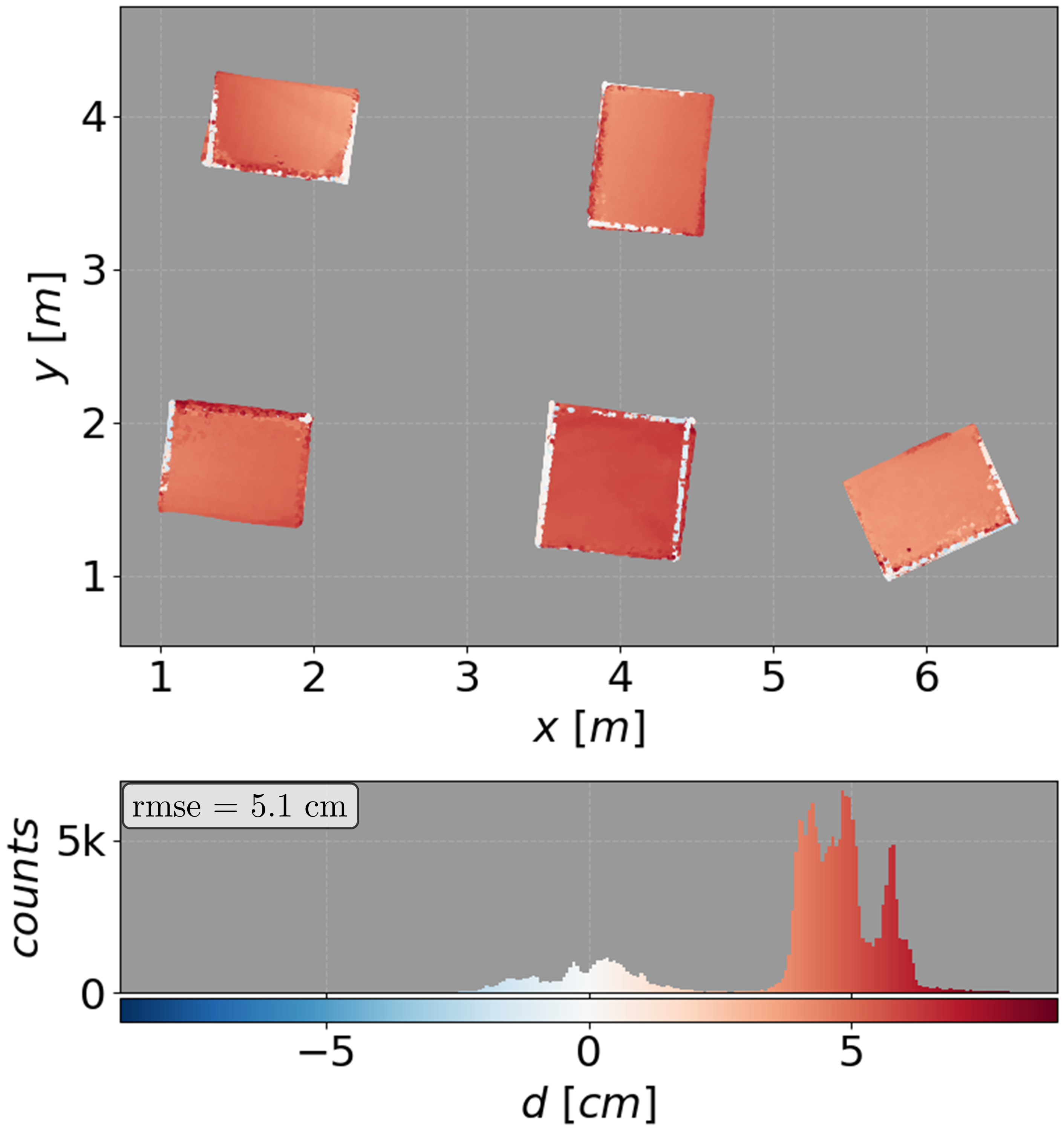}
        \caption{Calibration without reference scan}
        \label{subfig:bild2}
    \end{subfigure}
    \begin{subfigure}{0.32\linewidth}
        \centering
        \includegraphics[width=\linewidth]{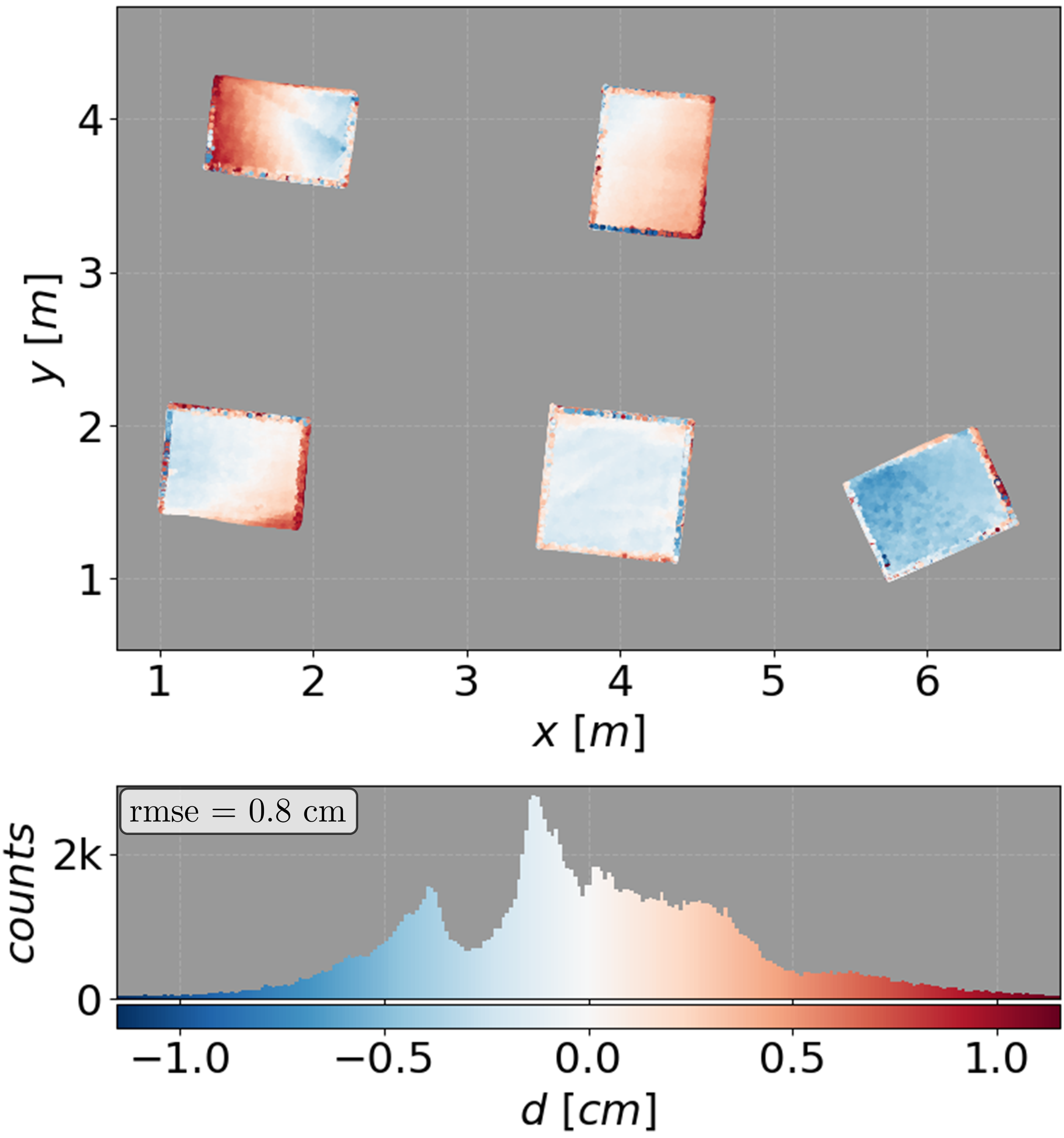}
        \caption{Calibration with reference scan}
        \label{subfig:bild3}
    \end{subfigure}
    \caption{Point cloud M3C2 distances $d$ between laser scanning system and reference TLS scans. The colors correspond to the M3C2 distances from negative (blue) to positive (red). The root-mean-squared errors of the M3C2 distances are shown in the upper left corner of the histogram plots.}
\end{figure*}

This section aims to investigate the influence of the reference scans on the parameter estimates. First, we estimate the calibration parameters with and without the reference scan with the calibration pipeline shown in figure \ref{fig:system_calibration_pipeline}. Afterward, the point clouds created with the initial guess parameters (a), estimated without reference (b), and with reference (c) are compared to the reference scan using the M3C2 metric. \\
%
%
Figure~\ref{subfig:bild1}-\ref{subfig:bild3} shows the point clouds using the initial and the estimated parameters. The colors refer to the M3C2 distances. The figure shows three comparisons: with initial parameters (fig.~\ref{subfig:bild1}), without reference scan (fig.~\ref{subfig:bild2}), and with reference scan (fig.~\ref{subfig:bild3}). The calibration parameters are summarized in table \ref{tab: calibration_parameter}. To make a comprehensible comparison between these three calibrations we compute the root-mean-squared-error of the M3C2 distances.
%
%
\begin{table}[!htb]
  \caption{Initial and estimated calibration parameters with and without reference scan for the left and right scanner}
  \begin{subtable}{\linewidth}
    \centering
    \caption{Left scanner}
    \begin{adjustbox}{max width=\linewidth}
    \begin{tabular}{c|cccccc}
                          & $t_x~[m]$ & $t_y~[m]$ & $t_z~[m]$ & $\alpha~[^\circ]$ & $\beta~[^\circ]$ & $\gamma~[^\circ]$ \\ \hline
        Initial           &  1.200  & -0.500  & 0.500 & -145.000 & 0.000  &  90.000 \\ 
        without TLS & 1.251  & -0.523  & 0.519 & -144.546 & -0.211 & 89.243 \\ 
        with TLS & 1.254 & -0.545 & 0.587 & -143.769 & -0.369  &  89.833 \\
    \end{tabular}
    \end{adjustbox}
  \end{subtable}
  \vspace{1em}
  \begin{subtable}{\linewidth}
    \centering
    \caption{Right scanner}
    \begin{adjustbox}{max width=\linewidth}
    \begin{tabular}{c|cccccc}
                          & $t_x~[m]$ & $t_y~[m]$ & $t_z~[m]$ & $\alpha~[^\circ]$ & $\beta~[^\circ]$ & $\gamma~[^\circ]$ \\ \hline
        Initial           &  1.200  & 0.650  & 0.500 & 145.000 & 0.000  &  90.000 \\ 
        without TLS &  1.217  &  0.659  &  0.526  & 144.491 & 1.312  & 89.456 \\ 
        with TLS & 1.259 & 0.656 & 0.585 & 144.396 & -0.346 & 91.705   \\ 
    \end{tabular}
    \end{adjustbox}
  \end{subtable}
  \label{tab: calibration_parameter}
\end{table}
The point cloud based on the initial calibration parameters and the histogram below show significant systematic deviations between -5\,cm and 9\,cm at an estimated M3C2 RMSE of 7.1\,cm,~fig.~\ref{subfig:bild1}. These large deviations come from inaccurately determined initial calibration parameters. The deviations can be explained by the multi-view scans, whereby the inaccurate calibration acts in different directions. \\
Figure \ref{subfig:bild2} shows the M3C2 distances using the parameters optimized without reference scan. The histogram spans from -2\,cm to +6\,cm at a RMSE of 5.1\,cm. Compared to figure~\ref{subfig:bild1} the systematic errors are reduced but they do not disappear. 
Especially the z translation of the calibration pronounces an offset of about 5\,cm, as also seen in table \ref{tab: calibration_parameter}. This can be explained by the fact, that with realistic trajectories of the robot (e.g. without flipping it upside down), a change of z-translation for both scanners does not influence the omnivariance values of the point cloud. The inclusion of a reference scan solves this problem.      
The calibration results with reference scan of figure \ref{subfig:bild3} show distances within an interval of $\pm$1\,cm, at a RMSE of 0.8\,cm. The remaining systematic deviations as they appear in figure~\ref{subfig:bild3} very likely result from the internal deformations of the robot, breaking the assumption of constant calibration parameters.
\subsection{Consistency of the calibration parameter} \label{subsec: param_consistency}
\begin{figure}[ht]
      \centering
      \includegraphics[width=1.0\linewidth]{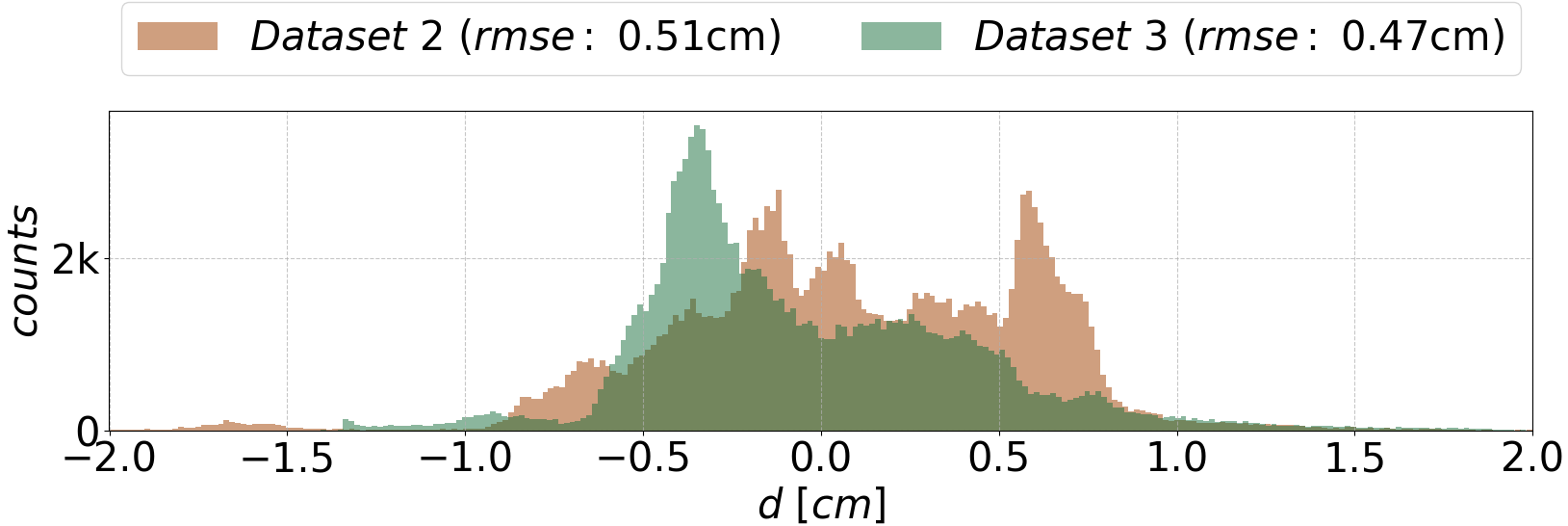}   
      \caption{M3C2 distances of reference and reconstructed point clouds from dataset 2 and 3, using the estimated calibration parameters of dataset 1 (with reference scan).}
      \label{fig:m3c2_2_3_cut}
\end{figure}

To evaluate the parameter consistency, we created point clouds from datasets 2 and 3, using the parameters estimated with dataset 1. The distances to the reference scan for both datasets are distributed within an interval of $\pm$1\,cm at a RMSE of $0.51$\,cm and $0.47$\,cm for datasets 2 and 3 respectively, see figure \ref{fig:m3c2_2_3_cut}. 
This magnitude of deviations shows the consistency of the calibration parameters of the first dataset. Again, remaining systematic deviations may result from robot deformations during the calibration procedure.
\subsection{Consistency between both scanners} \label{subsubsec: between}
For the task of precise crop reconstruction, the consistency between the left and the right scanner has a higher priority than the global accuracy of the points. We therefore analyse this consistency by calculating the distances between the clouds from both scanners.
\begin{table}[ht]
\centering
    \begin{tabular}{c|ccc}
             & \multicolumn{1}{c}{Dataset 1} & \multicolumn{1}{c}{Dataset 2} & Dataset 3 \\ \hline
    Initial  & \multicolumn{1}{c}{2.38}      & \multicolumn{1}{c}{2.47}      & 2.20      \\ 
    without TLS   & \multicolumn{1}{c}{1.82}      & \multicolumn{1}{c}{1.93}      & 1.66      \\ 
    with TLS & \multicolumn{1}{c}{1.51}      & \multicolumn{1}{c}{1.82}      & 1.53      \\ 
    \end{tabular}
    \caption{Root-mean-squared-errors of the M3C2 distances in centimeters between the point clouds of the left and right scanner.}
    \label{tab: m3c2_between}
\end{table}
Table \ref{tab: m3c2_between} summarizes the M3C2 RMSE between the two scanner point clouds, created by applying the calibration parameters of table \ref{tab: calibration_parameter}. The RMSE values with the initial parameters are in the range of 2.2 and -2.38\,cm, after calibration without reference scan the error decreases to 1.66 and 1.82\,cm. The smallest error of 1.51 and 1.82\,cm is achieved with the included reference scan. These results show that a higher consistency of the point clouds of both scanners is achieved by optimizing the calibration parameters as proposed in this paper. The results with and without reference scans show only small differences, indicating that the reference scan is more important for the absolute accuracy of the point cloud than for the internal consistency.
%
%
\section{CONCLUSIONS} \label{sec: conclusion}
%
%
In this work, we presented a calibration procedure for an agricultural field robot with multiple high-accuracy low-field-of-view, and low-range single-profile laser scanners. The goal of the calibration is to enable the creation of consistent point clouds of field crops, that have a very high resolution and accuracy. The method uses the scans from the different scanners while driving through a set of dedicated objects, as well as a static, terrestrial laser scan of the scene. It also needs accurate poses of the robot during the calibration, which we achieve by integrating total station measurements with IMU data in a factor graph-based estimation procedure.
Our results show that the accuracy of the resulting point cloud is within $\pm$1\,cm when compared with the static terrestrial laser scan as a reference. We also showed that the calibration parameters are transferable between datasets, leading to similar deviations from a ground truth point cloud of higher accuracy.

Nevertheless, we also observed remaining systematic deviations in our results, due to a non-static system calibration, caused by deformations of the robot enclosure. This requires a challenging kinematic system calibration along the robot trajectory which will be the focus of our future research.
%
%
                                  
\section*{ACKNOWLEDGMENT}
This work was funded by the Deutsche Forschungsgemeinschaft (DFG, German Research Foundation) under Germany’s Excellence Strategy–EXC 2070–390732324.

\bibliographystyle{IEEEtran}
\bibliography{references}{}

\begin{thebibliography}{10}
\providecommand{\url}[1]{#1}
\csname url@samestyle\endcsname
\providecommand{\newblock}{\relax}
\providecommand{\bibinfo}[2]{#2}
\providecommand{\BIBentrySTDinterwordspacing}{\spaceskip=0pt\relax}
\providecommand{\BIBentryALTinterwordstretchfactor}{4}
\providecommand{\BIBentryALTinterwordspacing}{\spaceskip=\fontdimen2\font plus
\BIBentryALTinterwordstretchfactor\fontdimen3\font minus
  \fontdimen4\font\relax}
\providecommand{\BIBforeignlanguage}[2]{{%
\expandafter\ifx\csname l@#1\endcsname\relax
\typeout{** WARNING: IEEEtran.bst: No hyphenation pattern has been}%
\typeout{** loaded for the language `#1'. Using the pattern for}%
\typeout{** the default language instead.}%
\else
\language=\csname l@#1\endcsname
\fi
#2}}
\providecommand{\BIBdecl}{\relax}
\BIBdecl

\bibitem{asseng2015rising}
S.~Asseng, F.~Ewert, P.~Martre, R.~P. R{\"o}tter, D.~B. Lobell, D.~Cammarano,
  B.~A. Kimball, M.~J. Ottman, G.~W. Wall, J.~W. White \emph{et~al.}, ``Rising
  temperatures reduce global wheat production,'' \emph{Nature climate change},
  vol.~5, no.~2, pp. 143--147, 2015.

\bibitem{asseng2019climate}
S.~Asseng, P.~Martre, A.~Maiorano, R.~P. R{\"o}tter, G.~J. O’Leary, G.~J.
  Fitzgerald, C.~Girousse, R.~Motzo, F.~Giunta, M.~A. Babar \emph{et~al.},
  ``Climate change impact and adaptation for wheat protein,'' \emph{Global
  change biology}, vol.~25, no.~1, pp. 155--173, 2019.

\bibitem{araus2014field}
J.~L. Araus and J.~E. Cairns, ``Field high-throughput phenotyping: the new crop
  breeding frontier,'' \emph{Trends in plant science}, vol.~19, no.~1, pp.
  52--61, 2014.

\bibitem{jay2015field}
S.~Jay, G.~Rabatel, X.~Hadoux, D.~Moura, and N.~Gorretta, ``In-field crop row
  phenotyping from 3d modeling performed using structure from motion,''
  \emph{Computers and Electronics in Agriculture}, vol. 110, pp. 70--77, 2015.

\bibitem{magistri2021towards}
F.~Magistri, N.~Chebrolu, J.~Behley, and C.~Stachniss, ``Towards in-field
  phenotyping exploiting differentiable rendering with self-consistency loss,''
  in \emph{2021 IEEE International Conference on Robotics and Automation
  (ICRA)}.\hskip 1em plus 0.5em minus 0.4em\relax IEEE, 2021, pp.
  13\,960--13\,966.

\bibitem{marks2022precise}
E.~Marks, F.~Magistri, and C.~Stachniss, ``Precise 3d reconstruction of plants
  from uav imagery combining bundle adjustment and template matching,'' in
  \emph{2022 International Conference on Robotics and Automation (ICRA)}.\hskip
  1em plus 0.5em minus 0.4em\relax IEEE, 2022, pp. 2259--2265.

\bibitem{weiss2011plant}
U.~Weiss and P.~Biber, ``Plant detection and mapping for agricultural robots
  using a 3d lidar sensor,'' \emph{Robotics and autonomous systems}, vol.~59,
  no.~5, pp. 265--273, 2011.

\bibitem{esser2023quality}
F.~Esser, L.~Klingbeil, L.~Zabawa, and H.~Kuhlmann, ``Quality analysis of a
  high-precision kinematic laser scanning system for the use of spatio-temporal
  plant and organ-level phenotyping in the field,'' \emph{Remote Sensing},
  vol.~15, no.~4, p. 1117, 2023.

\bibitem{qiu2019field}
Q.~Qiu, N.~Sun, H.~Bai, N.~Wang, Z.~Fan, Y.~Wang, Z.~Meng, B.~Li, and Y.~Cong,
  ``Field-based high-throughput phenotyping for maize plant using 3d lidar
  point cloud generated with a “phenomobile”,'' \emph{Frontiers in plant
  science}, vol.~10, p. 554, 2019.

\bibitem{esserfield}
F.~Esser, R.~A. Rosu, A.~Corneli{\ss}en, L.~Klingbeil, H.~Kuhlmann, and
  S.~Behnke, ``Field robot for high-throughput and high-resolution 3d plant
  phenotyping,'' \emph{IEEE Robotics \& Automation Magazine}, 2023.

\bibitem{kaartinen2012benchmarking}
H.~Kaartinen, J.~Hyypp{\"a}, A.~Kukko, A.~Jaakkola, and H.~Hyypp{\"a},
  ``Benchmarking the performance of mobile laser scanning systems using a
  permanent test field,'' \emph{Sensors}, vol.~12, no.~9, pp. 12\,814--12\,835,
  2012.

\bibitem{heinz2015development}
E.~Heinz, C.~Eling, M.~Wieland, L.~Klingbeil, and H.~Kuhlmann, ``Development,
  calibration and evaluation of a portable and direct georeferenced laser
  scanning system for kinematic 3d mapping,'' \emph{Journal of Applied
  Geodesy}, vol.~9, no.~4, pp. 227--243, 2015.

\bibitem{chan2013multi}
T.~O. Chan, D.~D. Lichti, and C.~L. Glennie, ``Multi-feature based boresight
  self-calibration of a terrestrial mobile mapping system,'' \emph{ISPRS
  journal of photogrammetry and remote sensing}, vol.~82, pp. 112--124, 2013.

\bibitem{rieger2010boresight}
P.~Rieger, N.~Studnicka, M.~Pfennigbauer, and G.~Zach, ``Boresight alignment
  method for mobile laser scanning systems,'' \emph{Journal of Applied
  Geodesy}, 2010.

\bibitem{yu2021automatic}
J.~Yu, X.~Lu, M.~Tian, T.~O. Chan, and C.~Chen, ``Automatic extrinsic
  self-calibration of mobile lidar systems based on planar and spherical
  features,'' \emph{Measurement Science and Technology}, vol.~32, no.~6, p.
  065107, 2021.

\bibitem{ravi2018bias}
R.~Ravi, Y.-J. Lin, M.~Elbahnasawy, T.~Shamseldin, and A.~Habib, ``Bias impact
  analysis and calibration of terrestrial mobile lidar system with several
  spinning multibeam laser scanners,'' \emph{IEEE Transactions on Geoscience
  and Remote Sensing}, vol.~56, no.~9, pp. 5261--5275, 2018.

\bibitem{maddern2012lost}
W.~Maddern, A.~Harrison, and P.~Newman, ``Lost in translation (and rotation):
  Rapid extrinsic calibration for 2d and 3d lidars,'' in \emph{2012 IEEE
  International Conference on Robotics and Automation}.\hskip 1em plus 0.5em
  minus 0.4em\relax IEEE, 2012, pp. 3096--3102.

\bibitem{hillemann2019automatic}
M.~Hillemann, M.~Weinmann, M.~S. Mueller, and B.~Jutzi, ``Automatic extrinsic
  self-calibration of mobile mapping systems based on geometric 3d features,''
  \emph{Remote Sensing}, vol.~11, no.~16, p. 1955, 2019.

\bibitem{elseberg2013automatic}
J.~Elseberg, D.~Borrmann, and A.~N{\"u}chter, ``Automatic and full calibration
  of mobile laser scanning systems,'' in \emph{Experimental Robotics: The 13th
  International Symposium on Experimental Robotics}.\hskip 1em plus 0.5em minus
  0.4em\relax Springer, 2013, pp. 907--917.

\bibitem{tombrink2023trajectory}
G.~Tombrink, A.~Dreier, L.~Klingbeil, and H.~Kuhlmann, ``Trajectory evaluation
  using repeated rail-bound measurements,'' \emph{Journal of Applied Geodesy},
  no.~0, 2023.

\bibitem{weinmann2017geometric}
M.~Weinmann, B.~Jutzi, and C.~Mallet, ``Geometric features and their relevance
  for 3d point cloud classification,'' \emph{ISPRS Annals of the
  Photogrammetry, Remote Sensing and Spatial Information Sciences}, vol.~4, pp.
  157--164, 2017.

\bibitem{dittrich2017analytical}
A.~Dittrich, M.~Weinmann, and S.~Hinz, ``Analytical and numerical
  investigations on the accuracy and robustness of geometric features extracted
  from 3d point cloud data,'' \emph{ISPRS journal of photogrammetry and remote
  sensing}, vol. 126, pp. 195--208, 2017.

\bibitem{dellaert2012factor}
F.~Dellaert, ``Factor graphs and gtsam: A hands-on introduction,''
  \emph{Georgia Institute of Technology, Tech. Rep}, vol.~2, p.~4, 2012.

\bibitem{forster2016manifold}
C.~Forster, L.~Carlone, F.~Dellaert, and D.~Scaramuzza, ``On-manifold
  preintegration for real-time visual--inertial odometry,'' \emph{IEEE
  Transactions on Robotics}, vol.~33, no.~1, pp. 1--21, 2016.

\bibitem{carlone2014eliminating}
L.~Carlone, Z.~Kira, C.~Beall, V.~Indelman, and F.~Dellaert, ``Eliminating
  conditionally independent sets in factor graphs: A unifying perspective based
  on smart factors,'' in \emph{2014 IEEE International Conference on Robotics
  and Automation (ICRA)}.\hskip 1em plus 0.5em minus 0.4em\relax IEEE, 2014,
  pp. 4290--4297.

\bibitem{lupton2011visual}
T.~Lupton and S.~Sukkarieh, ``Visual-inertial-aided navigation for high-dynamic
  motion in built environments without initial conditions,'' \emph{IEEE
  Transactions on Robotics}, vol.~28, no.~1, pp. 61--76, 2011.

\bibitem{kaess2012isam2}
M.~Kaess, H.~Johannsson, R.~Roberts, V.~Ila, J.~J. Leonard, and F.~Dellaert,
  ``isam2: Incremental smoothing and mapping using the bayes tree,'' \emph{The
  International Journal of Robotics Research}, vol.~31, no.~2, pp. 216--235,
  2012.

\bibitem{kaess2011isam2}
M.~Kaess, H.~Johannsson, R.~Roberts, V.~Ila, J.~Leonard, and F.~Dellaert,
  ``isam2: Incremental smoothing and mapping with fluid relinearization and
  incremental variable reordering,'' in \emph{2011 IEEE International
  Conference on Robotics and Automation}.\hskip 1em plus 0.5em minus
  0.4em\relax IEEE, 2011, pp. 3281--3288.

\bibitem{lague2013accurate}
D.~Lague, N.~Brodu, and J.~Leroux, ``Accurate 3d comparison of complex
  topography with terrestrial laser scanner: Application to the rangitikei
  canyon (nz),'' \emph{ISPRS journal of photogrammetry and remote sensing},
  vol.~82, pp. 10--26, 2013.

\end{thebibliography}

\end{document}